%% file: paper.tex
\documentclass[runningheads]{llncs}
\usepackage[T1]{fontenc}
\usepackage{graphicx}

\newcommand{\eg}{\emph{e.g.}}
\newcommand{\ie}{\emph{i.e.}}
\newcommand{\etal}{\emph{et~al.}}

\usepackage{tabularx}
\usepackage{caption}
\usepackage{subcaption}
\usepackage{cite}
\usepackage{url}
\usepackage{booktabs}
\usepackage[hidelinks]{hyperref}
\usepackage[export]{adjustbox}

\begin{document}
%
%\title{Learning X-ray Image Representations by Language Supervised Pretraining}
\title{RadTex: Learning Efficient Radiograph Representations from Text Reports}
%
%\titlerunning{Abbreviated paper title}
% If the paper title is too long for the running head, you can set
% an abbreviated paper title here
%
%\author{First Author\inst{1}\orcidID{0000-1111-2222-3333} \and
%Second Author\inst{2,3}\orcidID{1111-2222-3333-4444} \and
%Third Author\inst{3}\orcidID{2222--3333-4444-5555}}

\author{Keegan Quigley \inst{1} \and Miriam Cha \inst{1} \and Ruizhi Liao \inst{2} \and Geeticka Chauhan \inst{2} \and \\ Steven Horng \inst{3} \and Seth Berkowitz \inst{3} \and Polina Golland \inst{2}}
\authorrunning{Quigley et al.}

% \author{Anonymous}
% \authorrunning{Anonymous}

\institute{MIT Lincoln Laboratory, Lexington MA, USA \and
CSAIL, Massachusetts Institute of Technology, Cambridge, MA, USA \and
Beth Israel Deaconess Medical Center, Harvard Medical School, Boston, MA, USA}

%\institute{Princeton University, Princeton NJ 08544, USA \and
%Springer Heidelberg, Tiergartenstr. 17, 69121 Heidelberg, Germany
%\email{lncs@springer.com}\\
%\url{http://www.springer.com/gp/computer-science/lncs} \and
%ABC Institute, Rupert-Karls-University Heidelberg, Heidelberg, Germany\\
%\email{\{abc,lncs\}@uni-heidelberg.de}}
%
\maketitle              % typeset the header of the contribution
%

% \fancypagestyle{firststyle}
% {
%   \fancyhf{}
%   \fancyfoot[C]{\footnotesize Page \thepage\ of \pageref{LastPage}}
%   \renewcommand{\headrulewidth}{0pt} % removes horizontal header line
% }
\begin{abstract}
Automated analysis of chest radiography using deep learning has tremendous potential to enhance the clinical diagnosis of diseases in patients. However, deep learning models typically require large amounts of annotated data to achieve high performance -- often an obstacle to medical domain adaptation. In this paper, we build a data-efficient learning framework that utilizes radiology reports to improve medical image classification performance with limited labeled data (fewer than 1000 examples). Specifically, we examine image-captioning pretraining to learn high-quality medical image representations that train on fewer examples. Following joint pretraining of a convolutional encoder and transformer decoder, we transfer the learned encoder to various classification tasks. Averaged over 9 pathologies, we find that our model achieves higher classification performance than ImageNet-supervised and in-domain supervised pretraining when labeled training data is limited.

\keywords{Multimodal representation learning  \and Data efficiency.}
\end{abstract}
% \nnfootnote{\tiny{DISTRIBUTION STATEMENT A. Approved for public release. Distribution is unlimited. \newline\newline This material is based upon work supported by the Old Program 1 under Air Force Contract No. FA8702-15-D-0001. Any opinions, findings, conclusions or recommendations expressed in this material are those of the author(s) and do not necessarily reflect the views of the Old Program 1. \newline\newline \textcopyright Massachusetts Institute of Technology. \newline\newline Delivered to the U.S. Government with Unlimited Rights, as defined in DFARS Part 252.227-7013 or 7014 (Feb 2014). Notwithstanding any copyright notice, U.S. Government rights in this work are defined by DFARS 252.227-7013 or DFARS 252.227-7014 as detailed above. Use of this work other than as specifically authorized by the U.S. Government may violate any copyrights that exist in this work.}}

%
%
%

\input{intro}

\input{method}

\input{exp}
\input{conc}

\scriptsize
\input{ack}

% \clearpage

%
% ---- Bibliography ----
%
% BibTeX users should specify bibliography style 'splncs04'.
% References will then be sorted and formatted in the correct style.
%
\bibliographystyle{splncs04}
\bibliography{paper}
%
%\begin{thebibliography}{8}
%\bibitem{ref_article1}
%Author, F.: Article title. Journal \textbf{2}(5), 99--110 (2016)
%
%\bibitem{ref_lncs1}
%Author, F., Author, S.: Title of a proceedings paper. In: Editor,
%F., Editor, S. (eds.) CONFERENCE 2016, LNCS, vol. 9999, pp. 1--13.
%Springer, Heidelberg (2016). \doi{10.10007/1234567890}
%
%\bibitem{ref_book1}
%Author, F., Author, S., Author, T.: Book title. 2nd edn. Publisher,
%Location (1999)
%
%\bibitem{ref_proc1}
%Author, A.-B.: Contribution title. In: 9th International Proceedings
%on Proceedings, pp. 1--2. Publisher, Location (2010)
%
%\bibitem{ref_url1}
%LNCS Homepage, \url{http://www.springer.com/lncs}. Last accessed 4
%Oct 2017
%\end{thebibliography}
\end{document}

%% file: intro.tex
\section{Introduction}

\begin{figure}[t]
%\centering 
\includegraphics[width=1.0\textwidth, left]{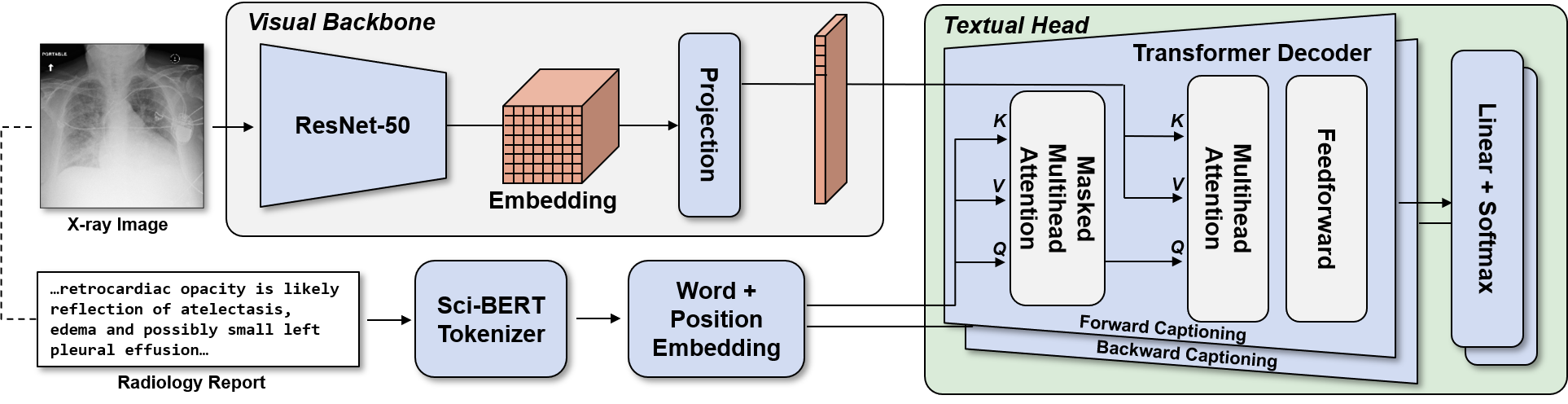} 
\caption{Overview of our RadTex framework. RadTex consists of a visual backbone based on ResNet-50 and textual head using transformer decoders, jointly trained from scratch on MIMIC-CXR v2.0 X-ray images and radiology reports. Element-wise sum and LayerNorm follow each layer in the transformer decoder, but are omitted from the visualization for simplicity.}
\label{fig:overview} 
\end{figure}  

Automated medical diagnostics from X-ray imagery have the potential to reduce strain on clinicians in hospital settings \cite{cdss_overview, davenport_ai}, and may even provide rapid insights into patient condition outside of the hospital \cite{ai_poc}. Deep learning (DL) and particularly convolutional neural networks (CNNs) have proven to be powerful tools for a wide range of computer vision tasks. Yet despite the advancement of CNNs, automated medical image assessment remains challenged by a lack of the massive expert-labeled datasets needed to successfully train such models. Extracting labels from existing clinical reports provides moderately sized labeled datasets for DL, but the process can be noisy and fails to provide the annotations needed for many high priority clinical tasks. Curating a medical dataset of comparable scale and quality to natural image datasets such as ImageNet (\ie, 1.4 million annotated images) is intractable because manual annotation of medical images by clinical experts is extremely time-consuming and expensive. In one study, for example, labeling pulmonary edema severity on 141 chest radiographs by 4 radiologists through a consensus process took approximately 25 man-hours \cite{deeplearning_pe}. Therefore, efficient learning from limited datasets on the scale of 100 labeled examples will be necessary as research models continue to be developed and implemented in the complex and high-stakes medical domain. 

%Use cases have therefore been limited to tasks where large-scale labels can reliably be extracted from existing reports. However, large-scale labels for many high-priority clinical tasks simply don't exist.

To address this need, there has been increasing interest in adopting pretraining weights, which can be transferred to downstream medical tasks with limited labeled data. A common method transfers model weights from supervised ImageNet-pretraining \cite{rethinking_pretraining, convirt}, yet there is a considerable gap between the natural image and medical image domains. In response to questions raised about the method's effectiveness in medical contexts \cite{raghu_2019}, some have attempted to bridge the domain gap by pretraining on existing large-scale labeled medical datasets \cite{taher_etal}. However, even this type of supervised, in-domain pretraining mostly relies on semantically sparse learning signals; labels are extracted through natural language processing (NLP) on accompanying text reports \cite{cxr14, irvin2019chexpert}, producing large-scale yet noisy annotations \cite{pneumothorax_detection} not ideal for building efficient representations for transfer learning.

Demonstrating alternatives to supervised pretraining, several recent studies leverage the unlabeled radiology reports for weak supervision. Clinical reports capture physicians' impressions in the form of unstructured text, yet contain rich semantic information. Prior work has capitalized on these natural annotations by using contrastive learning to encourage similar feature representations between image and text \cite{convirt, joint_chauhan, miccai_ruizhi}. While these methods attempt to learn representations from images and text jointly, we take a different approach, learning representations by predicting text \textit{from} images. The report generation task is an active area of research \cite{miura2020, alfarghaly, knee_reports}, and Wang~\etal~\cite{tienet} even demonstrated that pretrained LSTM captioning models improve downstream classification of medical images, but efficiency of learned representations for transfer is yet unknown.

%Such methods have also been extended to leverage additional labeling and common semantic patterns within reports through the use of max-margin and mutual information maximization objectives, respectively \cite{joint_chauhan, miccai_ruizhi}

% Intuitively, generating a full caption from an input image requires a more advanced understanding of the relationship between images and text compared to the ability to distinguish between matching and non-matching image-text pairs. 

%\textit{Chauhan}~\etal~\cite{joint_chauhan} applied a max-margin contrastive objective to encourage feature representations between true radiograph and radiology report pairs to be closer than randomly selected pairs. \textit{Zhang}~\etal~\cite{convirt} and \textit{Liao}~\etal~\cite{miccai_ruizhi} also used contrastive learning to maximize agreement between true radiograph and radiology report pairs. 

In this work, we aim to learn efficient radiograph representations from scratch that train on \textit{fewer} images. Recently, Desai \etal~\cite{virtex} introduced the VirTex framework showing that pretraining an image encoder via captioning leads to high-quality representations of natural images, which match or exceed those learned through ImageNet-supervised and contrastive pretraining, despite requiring fewer images for training. Inspired by the VirTex model's data-efficiency, we investigate radiograph representations learned from an image-captioning pretraining task and their potential for transfer to downstream medical image classification tasks. Specifically, we examine downstream performance under conditions of limited labeled data, comparing our model (RadTex) to other pretrained models. We find that (A) RadTex pretraining outperforms ImageNet and in-domain supervised pretraining methods when fewer than 1000 examples are available for training, (B) decreases in performance of just 0.05 AUC and 0.01 AUCPR are observed for RadTex when training data is reduced from tens of thousands of examples to 100 examples, and (C) pretraining RadTex on in-domain image-text datasets with at least 100K examples is necessary to build transferable and interpretable representations. 

% While others have explored image-to-text radiology report generation \cite{miura2020, alfarghaly, knee_reports}, to the best of our knowledge, this is the first time report generation has been used as a pretraining task for learning more efficient representations for downstream medical image classification.

% Our model, RadTex, consists of an image encoder and a text decoder jointly trained to generate a radiology report given an input radiograph. The learned image encoder is then transferred to downstream medical image deep learning tasks such as pathology- and edema severity-classification. 

% 1. Image-captioning pretraining outperfoms ImageNet and In-domain supervised pretraining methods when fewer than 1000 examples are available for clinical x-ray classification tasks.
% 2. Performance does not decrease from  
% 3 We attempt to find a size for a viable pretraining dataset for image-captioning

% The rest of this paper is structured as follows. Section 2 describes the overview of RadTex language supervised pretraining. In Section 3, we apply RadTex to edema severity classification and CheXpert pathology classification and compare its performance with image-only and ImageNet pretrained models with limited labeled data. In Section 4, we conclude the paper. 

%\begin{figure}[t]
%\centering 
%\includegraphics[width=.8\textwidth]{figs/intro_fig.pdf} 
%\caption{temp}
%\label{fig:highlevel} 
%\end{figure}  

%% file: method.tex
\section{Method}
We aim to use radiology reports to provide weak supervision for learning transferable radiograph representations. Compared to labels or segmentation annotations, radiology reports provide semantically dense information about the contents of a radiograph. Not only do reports note presence or absence of phenomena, but they also describe inter-relatedness between conditions and provide supporting evidence for diagnoses. The recent work of Desai~\etal~\cite{virtex} exploits the semantic density of text, showing that transferable representations of images can be learned from corresponding textual captions by jointly training an image encoder and text decoder. We adapt the VirTex model to the radiology domain, training an encoder-decoder architecture on MIMIC-CXR v2.0 X-rays and radiology reports \cite{MIMIC-CXR}. TieNet \cite{tienet} similarly applied image-captioning to the radiology domain, but in contrast, we train our model from scratch, use transformer decoders for captioning, and explore efficiency of learned representations. 

% The code is publicly available at X. Please refer to our website for more details about the datasets, architectures, and training procedures.

\subsection{Network Architecture}
Our model, RadTex, closely mirrors the architecture of VirTex with two main components: a visual backbone and textual head (encoder and decoder, respectively). Augmented X-rays are input to a visual backbone, consisting of a ResNet-50 convolutional network and a linear projection layer, which flattens embeddings to allow for decoder attention over visual features. Corresponding radiology reports are normalized, tokenized, and embedded using a learned token and positional embedding. After element-wise sum, layer normalization and dropout, these vector embeddings are passed to the textual head, which uses transformer decoders \cite{Attention} to predict masked tokens. The decoders use the context of visual features along with prior token predictions to predict the next token in a sequence. Like VirTex, we use bidirectional prediction of tokens during training. An overview of the RadTex model architecture is shown in Figure~\ref{fig:overview}. 

During transfer, the ResNet-50 is extracted from the visual backbone, and a fully connected linear layer is added as a decision head for supervised classification tasks. While we only report downstream classification results here, the linear decision head could be replaced with networks suited to other downstream tasks. We note that non-convolutional encoders, such as Vision Transformers\cite{ViT, Krishnan_2021}, could be incorporated into the visual backbone in the place of ResNet.

\subsection{Adapting VirTex to the Radiology Domain}
Numerous differences exist between the COCO image-text dataset \cite{mscoco} and MIMIC-CXR radiograph-report dataset. X-ray data are grayscale, and distinguishing content is present in small, local regions of the image. In contrast, COCO images are available in RGB, and context is gained by looking at wide regions of the image. We follow Xie~\etal~\cite{GrayscaleMedical} and alter the first layer of the standard ResNet-50 architecture used in VirTex to accept 1-channel images. We also adjust the size of the input images to 256$\times$256.

Radiology reports use highly technical language specific to the practitioner's domain to convey findings and contain multiple sections, including \textit{Impressions, Findings, Conclusion,} and \textit{Recommendation}. They also often compare findings to a baseline--either a healthy individual or a previous study of the patient--and  describe \textit{absence} as well as presence of phenomena (\eg, "No visible pneumothorax"). This is in stark contrast to COCO captions, which use simple language to describe only the important details present in the scene, and avoid speculation or comparison \cite{mscoco_captions}. To address some of these differences, we use a vocab list specific to the scientific domain, \texttt{\textit{scibert\_scivocab\_uncased}} \cite{beltagy-etal-2019-scibert}, take text only from the \textit{Findings} section of the report, and lengthen the maximum caption size that the model can accept to 170 tokens (95\textsuperscript{th} percentile of finding section length).

%% file: exp.tex
\section{Experimental Results}

In our experiments, we investigate RadTex's effectiveness in  learning visual representations during pretraining and its downstream transfer efficiency compared to other models with the same encoder architecture: randomly initialized ResNet-50, ImageNet-pretrained \cite{resnet}, and ChestX-ray14-pretrained \cite{cxr14, taher_etal} (CXR14).

\subsection{Datasets}

All examples used in RadTex training come from the MIMIC-CXR v2.0 dataset \cite{mimic_cxr_v2}, which includes a total of 247,425 pairs of frontal-view chest radiographs and their corresponding radiology reports. Two datasets that label pathologies in subsets of MIMIC-CXR were used for downstream study:\\
\textbf{Pathology9}~\cite{miccai_ruizhi}: Binary classification labels for pathology presence from \textit{MIMIC-CXR-JPG}\cite{MIMIC-CXR-JPG}, which applied the CheXpert NLP model \cite{irvin2019chexpert} to label 247,389 images in the MIMIC-CXR v2.0 dataset using their accompanying reports. We adhere to the official data split as used in Liao~\etal~\cite{miccai_ruizhi} and train models to classify each pathology individually as positive/negative. \\
\textbf{EdemaSeverity}~\cite{edema_severity}: 6,524 examples from MIMIC-CXR with pulmonary edema severity labels (0 to 3, increasing severity) extracted from the radiology reports using a regex model \cite{irvin2019chexpert, severity_grades}. Of these, 141 radiographs were examined by radiologists and consensus was reached on severity level (Fleiss' kappa: 0.42 individually, 0.97 consensus) \cite{deeplearning_pe}. We use regex-labeled examples during pretraining and consensus examples during transfer.

% \begin{itemize}
%     \item \textbf{Pathology9}~\cite{miccai_ruizhi}: Binary classification labels for pathology presence from \textit{MIMIC-CXR-JPG}\cite{MIMIC-CXR-JPG}, which applied the CheXpert NLP model \cite{irvin2019chexpert} to label 247,389 images in the MIMIC-CXR v2.0 dataset using their accompanying reports. We adhere to the official data split as used in Liao~\etal~\cite{miccai_ruizhi} and train models to classify each pathology individually as positive/negative.
%     \item \textbf{EdemaSeverity}~\cite{edema_severity}: 6,524 examples from MIMIC-CXR with pulmonary edema severity labels (0 to 3, increasing severity) extracted from the radiology reports using a regex model \cite{irvin2019chexpert, severity_grades}. Of these, 141 radiographs were examined by radiologists and consensus was reached on severity level (Fleiss' kappa: 0.42 individually, 0.97 consensus) \cite{deeplearning_pe}. We use regex-labeled examples during pretraining and consensus examples during transfer.
% \end{itemize}	

% During pretraining, we hold out all regex-labeled examples, and during transfer we use consensus examples as our test set.

\subsection{Training Details}
\label{section:training}
To study the efficiency of learned representations, we vary the number of examples in pretraining and downstream datasets and examine the model performance. For each phase of training, we convert images to 1-channel grayscale and perform random affine transformations, keeping the scale of the image constant. For ImageNet- and CXR14-pretrained models, we keep 3-channel inputs to match the pretrained architecture. 

% To handle class imbalance in Pathology9, we upsample the minority class (negative or positive) to match the number of majority examples. As a result, examples from the minority class would appear multiple times in the new training set, albeit with different augmentations. 

\begin{figure}[t]
    \centering
    \includegraphics[width=1.0\textwidth]{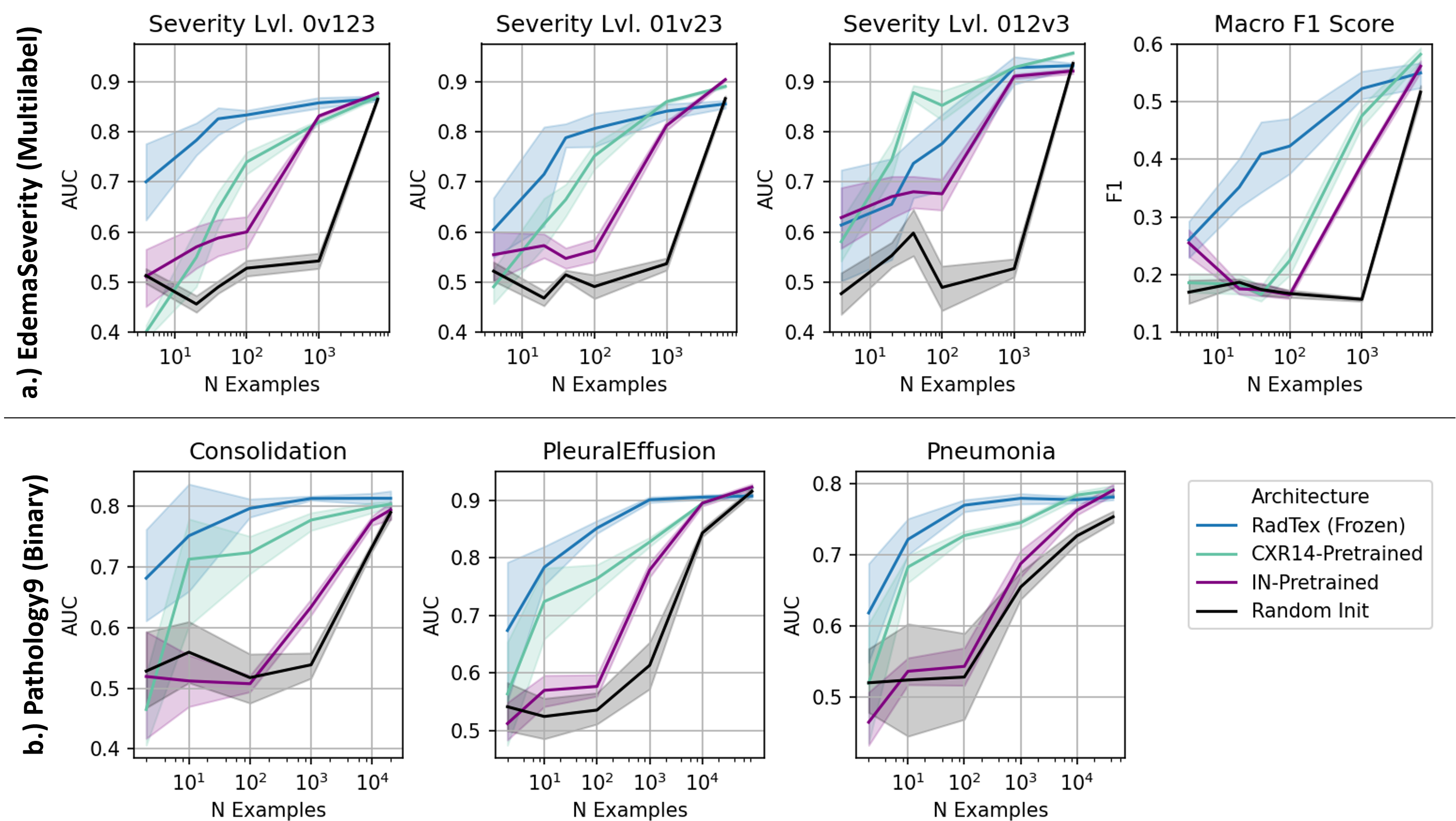}
        \caption{AUC with a varying amount of labeled training images ($N$) from a) EdemaSeverity and b) Pathology9. We compare frozen RadTex to other initializations, as unfrozen RadTex results were similar. Mean AUC from five trials and 95\% confidence interval is shown. Macro F1 score is reported for EdemaSeverity.} 
    \label{fig:chexpert_frozen}
\end{figure}

Pretraining was done separately for Pathology9 and EdemaSeverity to ensure that the test sets were not seen during pretraining. We pretrain using the same hyperparameters as Desai~\etal~\cite{virtex}, including SGD, LookAhead, and Weight decay. However, we train with a batch size of 128 images across 2 NVIDIA Volta V100 GPUs and do not use an early stopping criterion, but instead pretrain for a full 50 epochs. For pretraining and all downstream training, we use a linear learning rate warmup \cite{linear_warmup} for 5\% of epochs followed by cosine decay for the remaining 95\% of training. We pretrain our models and train all downstream models five times to estimate uncertainty in performance.

\begin{table}[t]
\scriptsize
\centering
\caption{Average AUC/AUCPR for Pathology9 with varying amounts of labeled downstream images. RadTex results use 100\% MIMIC pretraining.}

\label{table:chexpert_avg}
\begin{tabular}{@{}lcccccccc@{}}
\toprule
Downstream Examples &
  \multicolumn{2}{c}{\textbf{100\%}} &
  \multicolumn{2}{c}{\textbf{1000 Ex.}} &
  \multicolumn{2}{c}{\textbf{100 Ex.}} &
  \multicolumn{2}{c}{\textbf{10 Ex.}} \\ \midrule
\multicolumn{1}{l|}{} &
  \multicolumn{1}{l}{AUC} &
  \multicolumn{1}{l|}{AUCPR} &
  \multicolumn{1}{l}{AUC} &
  \multicolumn{1}{l|}{AUCPR} &
  \multicolumn{1}{l}{AUC} &
  \multicolumn{1}{l|}{AUCPR} &
  \multicolumn{1}{l}{AUC} &
  \multicolumn{1}{l}{AUCPR} \\ \midrule
\multicolumn{1}{l|}{Random Init} &
  0.770 &
  \multicolumn{1}{c|}{0.851} &
  0.628 &
  \multicolumn{1}{c|}{0.800} &
  0.536 &
  \multicolumn{1}{c|}{0.751} &
  0.549 &
  0.762 \\ \midrule
\multicolumn{1}{l|}{IN-Pretrained} &
  0.763 &
  \multicolumn{1}{c|}{0.849} &
  0.688 &
  \multicolumn{1}{c|}{0.820} &
  0.586 &
  \multicolumn{1}{c|}{0.770} &
  0.555 &
  0.761 \\ \midrule
\multicolumn{1}{l|}{CXR14-Pretrained} &
  \textbf{0.801} &
  \multicolumn{1}{c|}{\textbf{0.855}} &
  0.738 &
  \multicolumn{1}{c|}{0.833} &
  0.688 &
  \multicolumn{1}{c|}{0.816} &
  0.608 &
  0.797 \\ \midrule
\multicolumn{1}{l|}{RadTex Frozen} &
  0.785 &
  \multicolumn{1}{c|}{0.853} &
  \textbf{0.785} &
  \multicolumn{1}{c|}{\textbf{0.852}} &
  \textbf{0.752} &
  \multicolumn{1}{c|}{\textbf{0.844}} &
  \textbf{0.675} &
  0.817 \\ \midrule
\multicolumn{1}{l|}{RadTex Unfrozen} &
  0.762 &
  \multicolumn{1}{c|}{0.849} &
  0.757 &
  \multicolumn{1}{c|}{0.843} &
  0.734 &
  \multicolumn{1}{c|}{0.840} &
  0.670 &
  \textbf{0.822} \\ \bottomrule
\end{tabular}
\end{table}

During downstream training, we use binary cross entropy loss for Pathology9, and cross entropy loss for multi-label EdemaSeverity classification. For all Pathology9 and EdemaSeverity training, we use no weight decay and the same maximum learning rates (LR). In experiments, randomly initialized models ran for 50 epochs, with a max LR of $2\times10^{-1}$, while all other models were trained for 20 epochs. Through hyperparameter sweeps, the following maximum learning rates were found to be optimal: $2\times10^{-3}$ for IN-Pretrained and RadTex Unfrozen, and $2\times10^{-2}$ for CXR14-pretrained and RadTex Frozen. 

%IN-Pretrained and RadTex Unfrozen were fine-tuned for 20 epochs, with a max LR of , and RadTex Frozen and CXR14-pretrained (fine-tuned) models ran 20 epochs with a max LR of  for training the decision head only. Hyperparameter sweeps found these to be the optimal learning rates for downstream training.

\subsection{Training with Fewer Labeled Images in Downstream Tasks}

Figure~\ref{fig:chexpert_frozen} compares area under the receiver operating characteristic curve (AUC) values for language-supervised RadTex models, CXR14-pretrained models from \cite{taher_etal}, ImageNet-pretrained models, and randomly initialized image-only models. We train all models on randomly sampled portions of the downstream training dataset to investigate the relationship between labeled dataset size and performance for each model. Notice that for both EdemaSeverity and Pathology9, RadTex trains representations that transfer to downstream tasks with much more efficiency than other models. When 1000 or fewer examples are available, RadTex matches or outperforms ImageNet- and CXR14-pretrained models. 

% We find that following RadTex pretraining, high quality visual models can be learned from relatively few training examples. 

Table~\ref{table:chexpert_avg} presents averaged AUC and area under the precision-recall curve (AUCPR) results across Pathology9. These findings suggest that RadTex is a promising approach to overcoming the challenge of limited labeled data, commonly found in the medical domain. Crucially, we found that RadTex Frozen performance on 100 labeled examples (AUC/AUCPR: 0.752/0.844) was nearly as good as the top performing model with 100\% labeled data (0.801/0.855),  where $\sim$100 is a reasonably sized dataset to ask physicians to annotate. 

As shown in both Figure~\ref{fig:chexpert_frozen} and Table~\ref{table:chexpert_avg}, all models perform similarly when all downstream data is used (100\% downstream). This is not unexpected--all classification models use the exact same architecture (ResNet-50 with decision head), and thus have the same predictive capacity. Given enough training data and epochs, we would expect performance to converge. However, when training data is limited, the RadTex initialization of models becomes advantageous.

\begin{figure}[t]
\centering 
\includegraphics[width=1\textwidth, left]{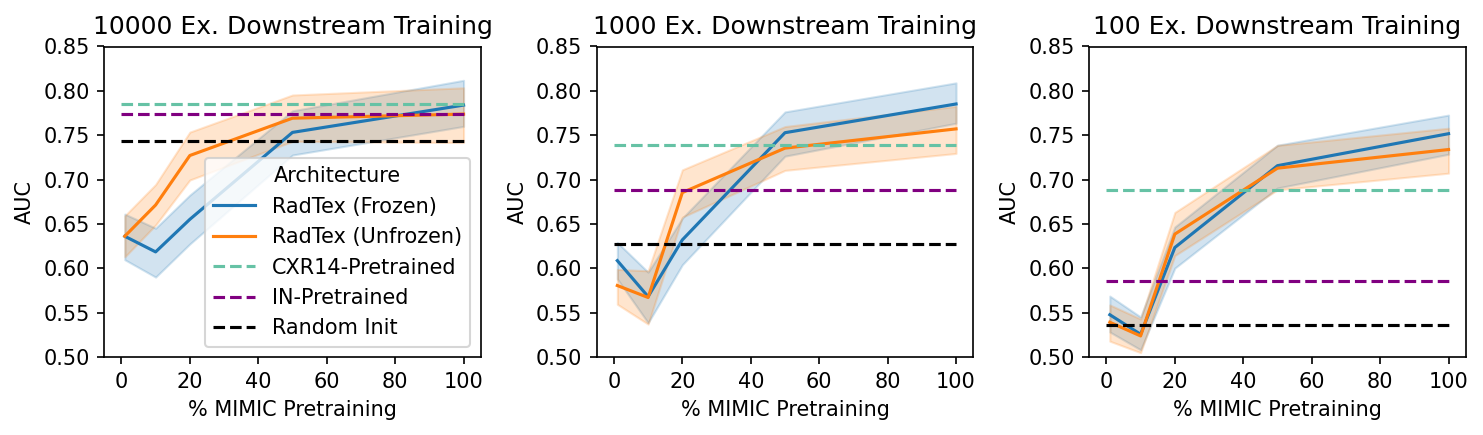} 
\caption{Averaged Pathology9 AUCs after training on 10K, 1K and 100 downstream examples vs. pretraining dataset (MIMIC-CXR) size.}
\label{fig:pretraining} 
\end{figure}  

% Means and 95\% confidence intervals are shown for each RadTex AUC.

% Comparison is made to other ResNet-50 initializations (dashed, no MIMIC pretraining).

\subsection{Pretrained Representation Quality}
While it is intuitive that pretraining with more examples would produce better visual embeddings and representations, we wanted to understand whether the size of the MIMIC-CXR corpus ($\sim$250K frontal X-ray/report pairs) was sufficient for the captioning pretraining task. In Figure~\ref{fig:pretraining}, we capture the effect of reducing the size of the pretraining set by transferring models to 10K, 1K, and 100 example downstream training on Pathology9. We observe that pretraining with a dataset size 50\% of the MIMIC corpus degrades downstream AUCs only slightly compared to the full pretraining set (100 example Pathology-9 drops by 0.03 AUC), yet pretraining with smaller datasets yields significantly worse performance, as the model seems unable to "recover" from a poor initialization. Noting that 50\% MIMIC pretraining ($\sim$125K examples) represents an inflection point in RadTex downstream performance in Fig~\ref{fig:pretraining}, and that VirTex similarly used a pretraining dataset with 118K examples, we recommend that a pretraining dataset of at least 100K examples be available for those looking to apply RadTex to additional radiographic modalities or imaged regions. 

% With additional training or a higher learning rate, we might expect the limited-pretraining RadTex unfrozen models to match the performance of random initialization, but benefits of language-supervised pretraining may not be realized.

\subsection{Proxy Task: Generating Radiology Reports}
\begin{figure}[t]
\centering 
\includegraphics[width=1\textwidth, left]{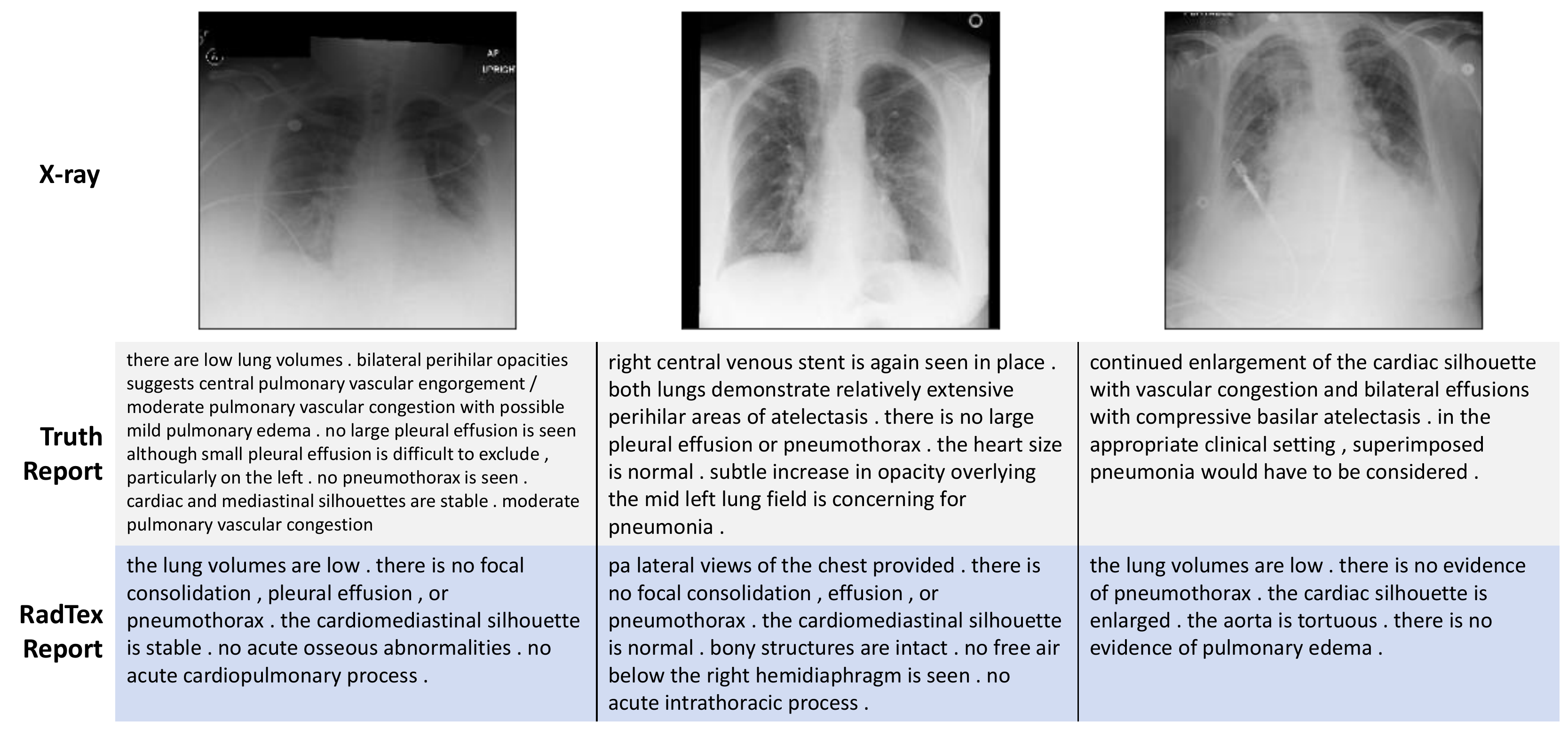} 
\caption{RadTex-generated reports (\textit{Findings}) compared to the originals.}
\label{fig:captioning} 
\end{figure}

Generating high-quality captions is not the goal of the VirTex or RadTex models, but the task provides interpretability and insight into the representations learned during pretraining. We experimented with radiology report generation in RadTex on the Pathology9 test set after pretraining on MIMIC-CXR. Following 200 epochs pretraining, we apply a beam search algorithm (2 beams) to predict tokens sequentially. As shown in Figure~\ref{fig:captioning}, RadTex-generated reports for three random radiographs from the test set show some agreement with the \textit{Findings} section of the corresponding report. However, while radiology reports build up evidence for diagnoses, these generated reports offer little supporting evidence, and occasionally even refer to information beyond what could be obtained from the image (\eg, "PA lateral views of the chest provided"). Nonetheless, similarities in pathology mentions between generated and written reports suggest that the model is doing more than mimicking the language of radiologists. 

%Captioning performance degraded for RadTex models pretrained with less than 100\% MIMIC data, indicating a drop in interpretability as well as transfer potential.

%% file: conc.tex
\section{Conclusions}
We present a model, RadTex, which leverages the semantic density of radiology reports to build efficiently transferable radiograph representations. We focus our analysis of the model on the effects of data availability, yielding insights into performance on expert-annotated datasets of reasonable curation size according to our clinical collaborators. By adopting radiology report generation as language-supervised pretraining, we observe that the learned image encoder can produce high-quality chest X-ray representations with as few as 100 labeled images, potentially reducing the labeling burden on radiologists for developing deep learning models. Results on various classification tasks suggest that RadTex is a superior approach to supervised pretraining for data-efficient transfer and interpretability. Additionally, our results suggest that pretraining on a large image-text corpus is important to achieving RadTex's competitive advantage.

%% file: ack.tex
\subsubsection*{Acknowledgements}
This work was supported in part by MIT Lincoln Laboratory, US Air Force, NIH NIBIB NAC P41EB015902, Wistron, IBM Watson, MIT Deshpande Center, and MIT J-Clinic. 

DISTRIBUTION STATEMENT A. Approved for public release. Distribution is unlimited. This material is based upon work supported by the Old Program 1 under Air Force Contract No. FA8702-15-D-0001. Any opinions, findings, conclusions or recommendations expressed in this material are those of the author(s) and do not necessarily reflect the views of the Old Program 1. \textcopyright Massachusetts Institute of Technology. Delivered to the U.S. Government with Unlimited Rights, as defined in DFARS Part 252.227-7013 or 7014 (Feb 2014). Notwithstanding any copyright notice, U.S. Government rights in this work are defined by DFARS 252.227-7013 or DFARS 252.227-7014 as detailed above. Use of this work other than as specifically authorized by the U.S. Government may violate any copyrights that exist in this work.